\documentclass[11pt,a4paper]{article}
\usepackage[hyperref]{acl2020}
\usepackage{times}
\usepackage{latexsym}

\usepackage{graphicx}
\usepackage{url}
\usepackage{amsmath}
\usepackage{bbm}
\usepackage{multicol}
\usepackage{graphicx}
\usepackage{booktabs}
\usepackage{subcaption}
\usepackage{microtype}
\aclfinalcopy % Uncomment this line for the final submission
\title{Adaptive Transformers for Learning Multimodal Representations}

\author{Prajjwal Bhargava \\
  \texttt{prajjwalgo@gmail.com}}

\date{}

\begin{document}
\maketitle
\begin{abstract}
The usage of transformers has grown from learning about language semantics to forming meaningful visiolinguistic representations. These architectures are often over-parametrized, requiring large amounts of computation. In this work, we extend adaptive approaches to learn more about model interpretability and computational efficiency. Specifically, we study attention spans, sparse, and structured dropout methods to help understand how their attention mechanism extends for vision and language tasks. We further show that these approaches can help us learn more about how the network perceives the complexity of input sequences, sparsity preferences for different modalities, and other related phenomena.
\end{abstract}

\section{Introduction}

% Learning about vision and language is crucial for embodied AI agents to solve complex tasks. Language models have proven to be useful in understanding language semantics, and its generation \cite{radford2019language}. Transformers \cite{vaswani2017attention} have been applied to a wide variety of language tasks and have achieved remarkable performance on most of them. BERT \cite{devlin2018bert} is a bi-directional language model that has proven to work well for transfer learning and fine-tuning after performing extensive pre-training on a large language corpus. However, dealing with a single modality is not sufficient to solve complex tasks that require understanding language with a visual context.
Learning richer representations from visual and text data is a central task to solve multi-modal learning. Attention-based methods have proven to be very useful in learning long term dependencies and forming richer representations of the input sequences. Numerous approaches \cite{lu2019vilbert, su2019vl, li2019visualbert, chen2019uniter} have been proposed for learning visiolinguistic representations with transformers. Although these approaches have provided us with significant improvement on various benchmarks (language and visiolinguistic), the
architectures used are over-parameterized require extensive training lasting for several weeks using multiple objectives to form a generalized representation of the task to be addressed, which is then followed by fine-tuning on a downstream task. This workflow has become a concerning problem. It results in deep learning methodologies being inaccessible and increased carbon footprints \cite{strubell2019energy}. In this work, we specifically explore adaptive methods. We refer to Adaptive mechanisms as those methods that change their behavior during training/run time and adapt stochastically to the environment based on data heuristics (parameters) learned by encountering samples from the same data distribution optimized by an objective function. Alternative approaches such as pruning, distillation \cite{hinton2015distilling} and quantization are rigid to some extent and induce some form of permanent modifications to the model. Adaptive methods enforce the network to learn parameters such that their behavior changes as per the complexity of the input sequence as perceived by the neural network. The code to reproduce the results  in this work is publicly available at this link\footnote{\url{https://github.com/prajjwal1/adaptive_transformer}}.

Current self-attention approaches assume that the attention span of a head is invariant to the complexity of an input sequence. Attention heads can learn their optimal context size \cite{sukhbaatar2019adaptive}, which results in a reduction of FLOPS. When an optimal attention span is learned, the amount of attention given to a particular input sequence by an attention head is determined by its context size. We show that the context size varies with the emergent complexity of the sequence, and spans can help us understand how much sensitive a layer is to an input sequence.

Training models with a quarter of a million parameters are not feasible and practical for most users. One effective way to facilitate neural network scaling is by making the weights of the network sparse. This configuration allows us to perform faster training of deeper networks with relatively less compute. To make attention distributions sparse, we use $\alpha$ entmax \cite{correia2019adaptively} to obtain probability distribution of weights. Normalized exponential functions like softmax cannot assign a zero attention weight. This property enforces the context vector to stay dense, resulting in non-relevant sequences to be considered even though the network has discarded them by putting a deficient weight. Adaptive sparsity can make an attention head to learn richer distributions by oscillating the behavior of distribution to stay between softmax and sparsemax. We show that this behavior can help us understand preferences for the density of attention weight distribution and how it varies amongst each head about different modality.

We also study a form of regularization method called Layerdrop ~\cite{fan2019reducing} to understand its regularization impact for multi-modal features. If the network can learn to drop identical layers (\emph{Data Driven} pruning), then it can be regarded as an adaptive depth mechanism.  We specifically use the \emph{Every other} pruning method where the user specifies the drop rate because it offers maximal gains as suggested compared to its counterpart pruning methods. This method has proven to be effective in reducing the number of parameters and pruning layers during inference.

The contribution of this work is as follows:
\begin{itemize}
    \item The adaptive approaches have only been tested with linguistic features only. We extend these approaches to study how do they align to capture complex relationships between different modalities. We also study the effects of aligning these approaches to understand their compatibility through ablation analysis.
    \item We perform interpretability analysis to learn how these approaches can enhance our understanding of attention behavior and adaptive approaches. 
    \item We provide experimental results on the recent adaptive approaches for the multi-modal input sequences.
\end{itemize}

% \section{Related Work}

% Transformers were initially used for learning textual representation. BERT \cite{devlin2018bert} was initially pre-trained on the Masked Language Model objective and achieved significantly better results on downstream tasks compared to approaches that solely relied on learning the target downstream task. Recently, \cite{cordonnier2019relationship} showed that self-attention layers could perform convolution and also proved that a self-attention layer is as expressive as a convolution layer. ViLBERT \cite{lu2019vilbert} and LXMERT \cite{tan2019lxmert} extends BERT by employing a two-stream architecture to process both visual and textual sequences which are made use of by coattentional transformer layers. VL BERT \cite{su2019vl} used a unified single-stream architecture for deriving generic representations for various visual-linguistic tasks. UNITER \cite{chen2019uniter} uses conditional masking i.e., masking one modality at a time during training along with similar pre-training methods to learn image-text relationships.

\section{Background}

\subsection{LXMERT}
We use LXMERT \cite{tan2019lxmert} as the baseline architecture. The adaptive approaches can be combined with any other self-attention mechanism based transformer. LXMERT uses self and cross attention layers to jointly attend to image and text inputs (input sequence). Specifically, it takes a word-level sentence and object-level image embeddings. The encoder consists of three main components: language (9 layers) and visual (5 layers) encoder (single-modality) to form textual and image representations and cross-modality encoder (5 layers) to jointly attend to both these representations. Cross attention is responsible for forming the mapping between ROI features and textual representations. Since the architecture used is identical, we refer the readers to \cite{tan2019lxmert} for a detailed description of pre-training strategies. The network used has been pre-trained on four objectives: Masked Cross Modality LM, Masked Object Prediction, Cross Modality Matching, and Image Question Answering. Faster RCNN \cite{NIPS2015_5638} is used to extract ROI features from the input images.
% \begin{itemize}
%     \item 
%     \item 
%     \item C
%     \item Image Question Answering
% \end{itemize}
% We use the pretrained weights provided by \cite{tan2019lxmert} followed by fine-tuning, which is performed with the mentioned adaptive approaches. 
% We initially perform masking of words with probability set as 0.15. The network has to predict the masked word with the help of the image. In this case, the network is made to learn about the cross-modality relationship by relying upon vision modality to solve the language-related task. 

% \subsubsection{}
% Similar to the previous task, we mask ROI features with zeros to eliminate objects in the image. The network is made to predict the label of the missing object from the remaining visual context and language sequence.

% \subsubsection{}
% Sentences are replaced with a probability of 0.5, and a classifier is trained to predict if the image is related to the text sequence or not. This helps the network learn about what kind of questions it should expect regarding a visual context. 

% \subsubsection{}
% In this task, we perform the standard task of QA. The network is made to predict to answers to the questions. In this case, the data is used as it is, no replacements of any sequence is performed. 

\begin{figure*}[t]
\centering
 % \begin{}{lllll}
    \includegraphics[width=.3\linewidth]{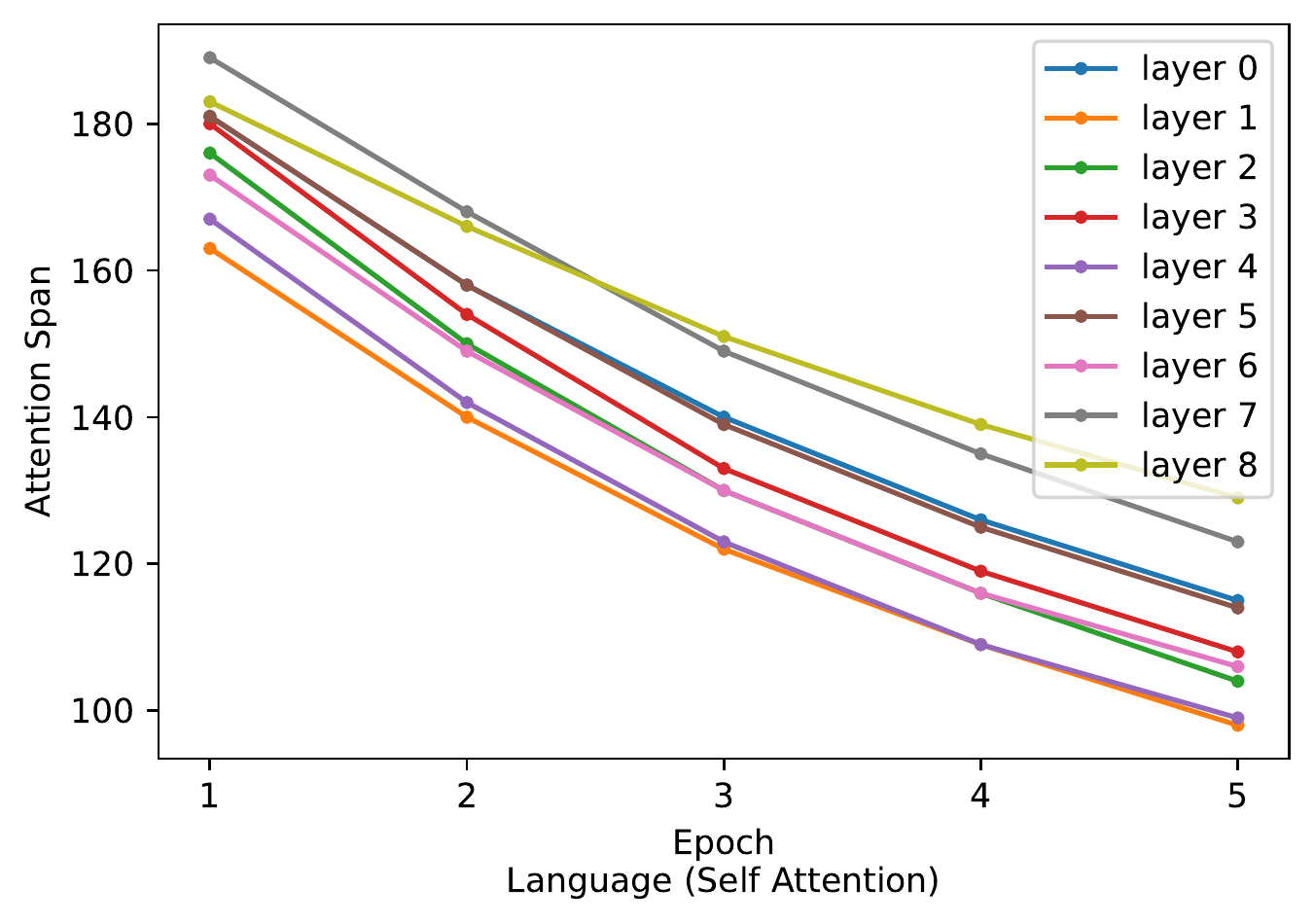}\quad
    \includegraphics[width=.3\linewidth]{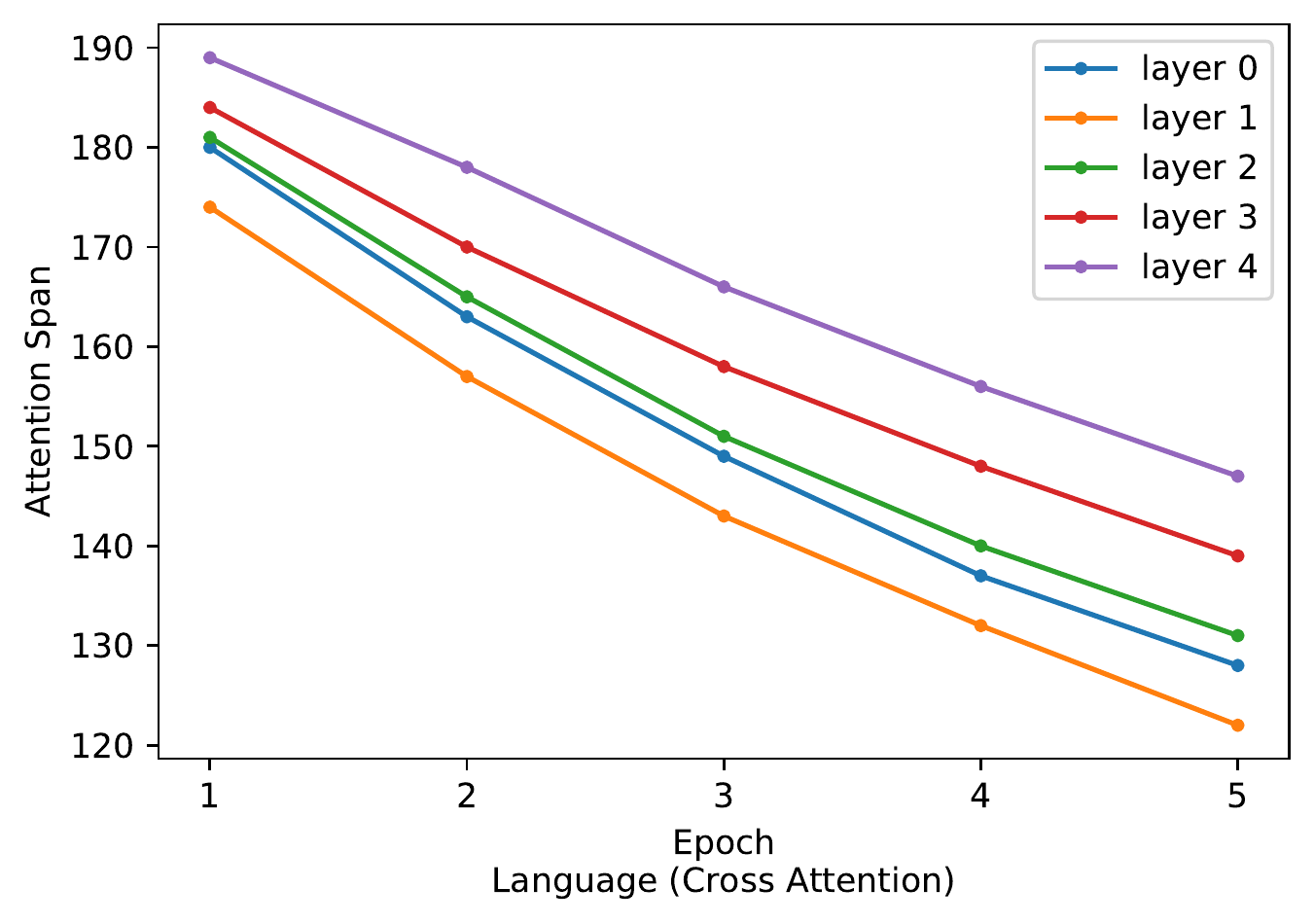}\quad
    \includegraphics[width=.3\linewidth]{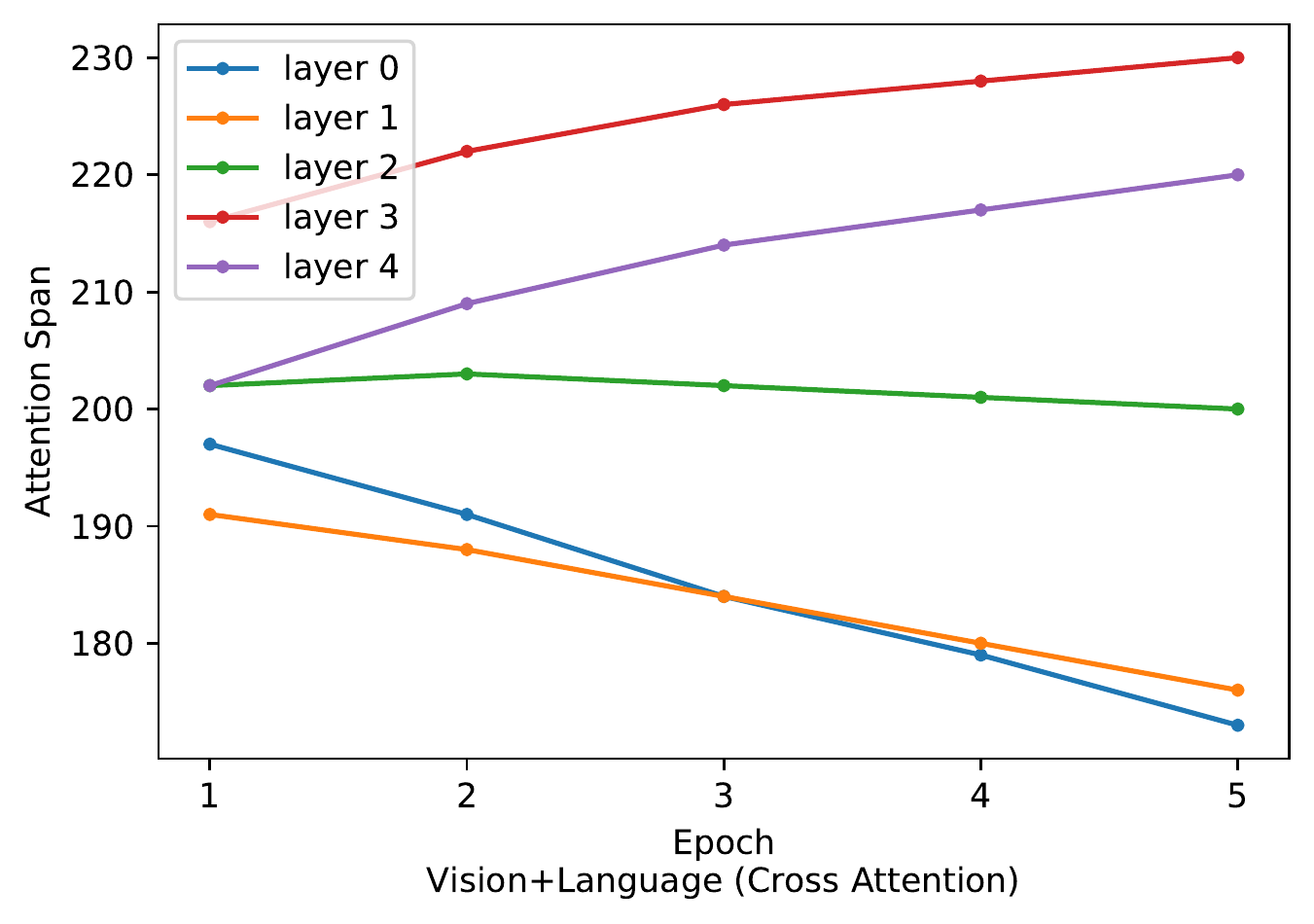} 
    \medskip
    \includegraphics[width=.3\linewidth]{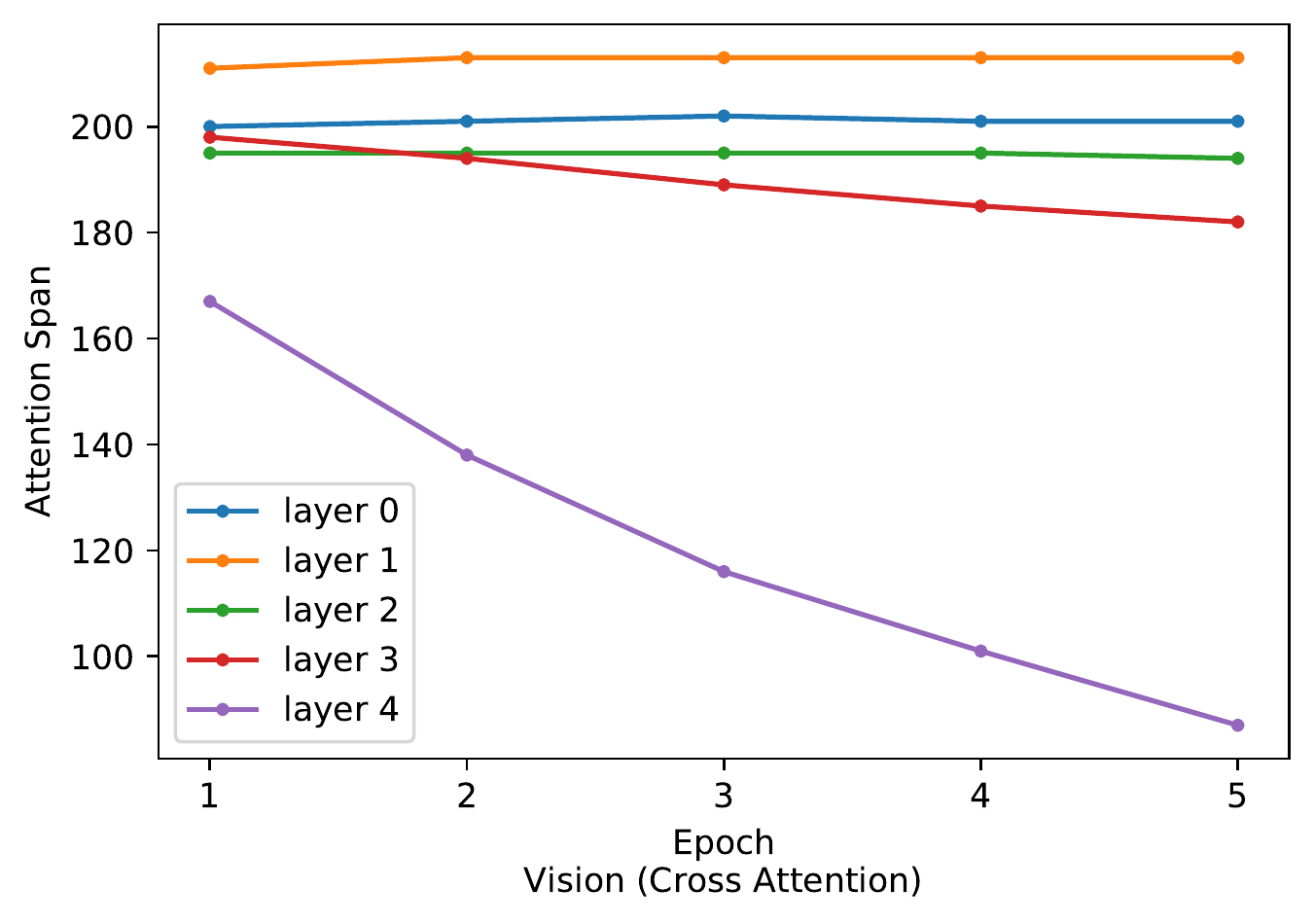}\quad
    \includegraphics[width=.3\linewidth]{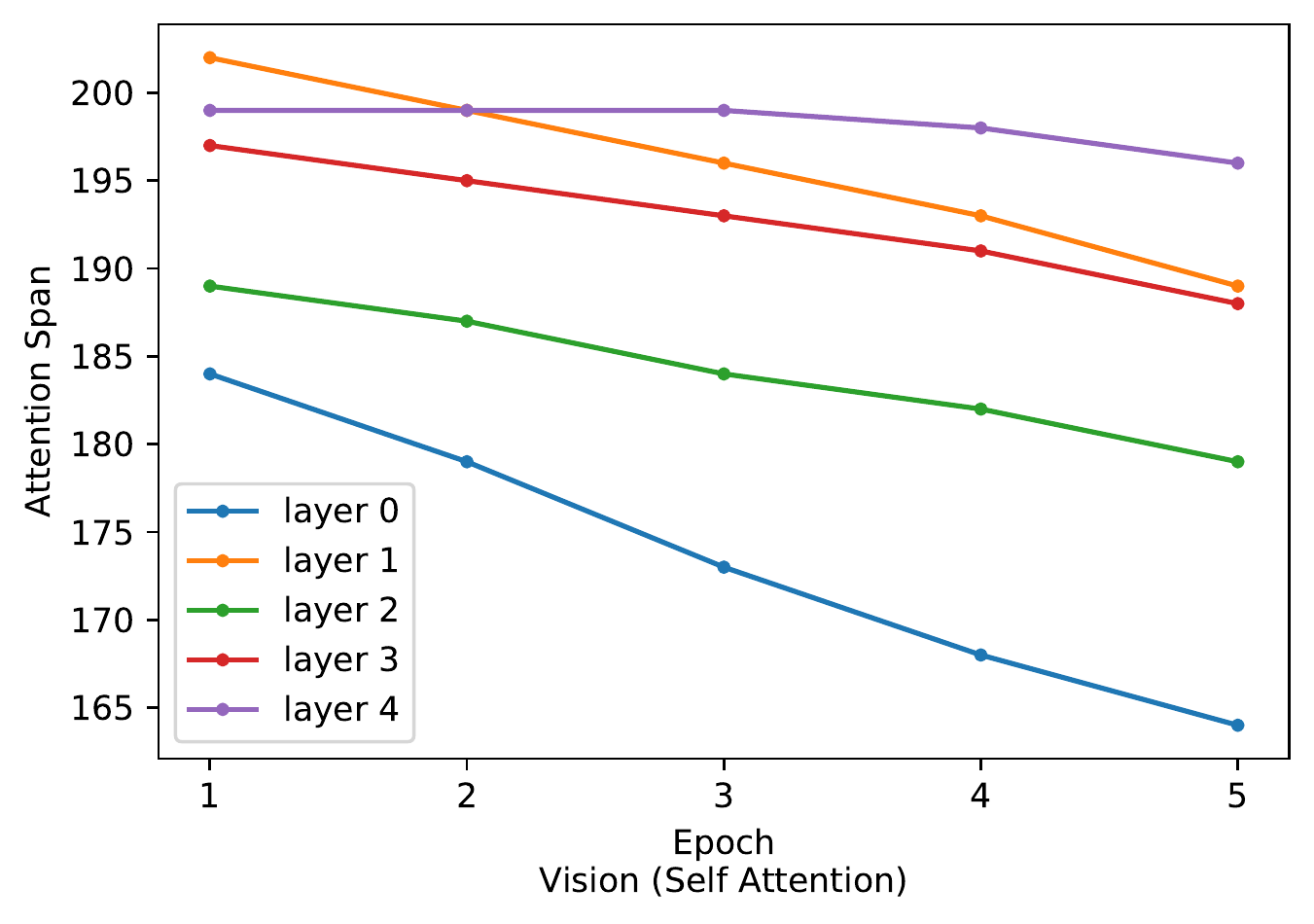}
 % \end{tabular}
  \vspace{-4mm}
  \caption{Variation of adaptive spans in different attention layers (single and cross-modality) as the training progresses. Accuracy on the local-validation set is reported per epoch. The maximum adaptive span limit was set to 1024}
\label{fig:span_results}
\end{figure*}

\subsection{Adaptive Attention Span}
Unlike dynamic attention, which assumes that all attention heads require the same amount of span, learning an optimal attention span enables the gathering of information as per the context size determined by the attention head. A max upper bound span limit is enforced on each head, which helps reduce computation and memory requirements. As proposed in \cite{sukhbaatar2019adaptive}, different heads emphasize on different context depending upon the task it is addressing. We explicitly show that these spans vary significantly based on the complexity of the task.
We use the same masking function with minor modification:
\begin{equation}
m_{z}(x)=\min \left[\max \left[\frac{1}{R}(R+z-x), 0\right], 1\right]
\end{equation}
Here, $z$ acts as a model's parameter. We initialize it with kaiming normal \cite{he2015delving} distribution. $m_{z}$ is coupled with the attention weights. Hyperparameter $R$ helps in controlling the softness of this attention distribution.

The attention head compute the similarities between current token $t$ and past token $r$ in the span $[t-S, t)$ as:
\begin{equation}
    s_{tr} = x_{t}^{T}Q^{T}(Kx_{r}+P_{t-r})
\end{equation}
where $K$, $Q$ and $P_{t-r}$ denote key, query vectors, and position embedding respectively. In the standard setting, attention weight distribution is obtained by applying softmax on the similarity vector.
\begin{equation}\label{eq: attention_softmax}
    A_{tr} = softmax(s_{tr})
\end{equation}
The attention weights from \autoref{eq: attention_softmax} are then processed by the masking function as:
\begin{equation}
    A_{tr} = \dfrac{m_z(t-r)exp(s_{tr})}{\sum\limits_{q=t-S}^{t-1}m_z(t-q)exp(s_{tr})}
\end{equation}

The masking function is a non-increasing function that applies a transformation to the input values of attention scores to keep them in range of $[0,1]$.
The parameters of $m_{z}$ are updated with model parameters to learn the optimal span.

\subsection{Adaptive Sparse Attention}
In order to make attention weights sparse, we use $\alpha$ entmax as proposed in ~\cite{correia2019adaptively}. Specifically, softmax is replaced with $\alpha$ entmax to compute attention weights given attention scores in \autoref{eq: attention_softmax}.
\begin{equation}
    \operatorname{Att}(Q, K, V)=\pi\left(\frac{Q K^{\top}}{\sqrt{d}}\right) V 
\end{equation}
\begin{equation}
    \pi(Z)_{i j}=\alpha \text { -entmax }\left(z_{i}\right)_{j} \\
\end{equation}
$\alpha$ plays a crucial role in determining the behavior of an attention head. If $\alpha > 1$, the weight distribution would move away from softmax's dense representation towards sparse mappings as its curvature changes. For $\alpha = 2$, we obtain complete sparse mappings. The value of alpha oscillates between 1 and 2.  It is set as a network parameter, which is jointly optimized in the training process. Different values of $\alpha$ will govern the behavior of the attention head. 

\subsection{LayerDrop}
Layerdrop ~\cite{fan2019reducing} is a method to reduce the depth of the transformer in a controlled manner. This method drops the identical sub-layers in the transformer determined by a pruning strategy. We follow the \emph{Every Other} strategy, which drops the layer as specified by a drop rate. It has been noted that this pruning strategy works well as compared to \emph{Search on Valid} and \emph{Data Driven} pruning strategies. 
Let $N$ denote the total number of layers in the network. Setting $p=1$ implies that we are dropping one layer out of all the layers assigned for a modality. The number of remaining layers becomes $N-p$. Although the network will consist of an equivalent amount of parameters as that of $N$ layers, all the operations will be carried out equivalent to operations in $N-p$ layers. This strategy allows us to prune layers during inference time.

\section{Experimental Setup}

\paragraph{Visual Question Answering} 
To solve the VQA task, given an image and a question related to it, the network is supposed to predict the right answer from the given set of answer choices. We performed all the experimentation on the VQA 2.0 dataset \cite{antol2015vqa}. The dataset consists of three sets with a train set containing 83k images and 444k questions, a validation set containing 41k images and 214k questions, and a test set containing 81k images and 448k questions. In this case, the network is asked to predict an answer from 3129 answer choices for a particular question. 

\paragraph{Implementation}
We use the pre-trained weights provided by ~\cite{tan2019lxmert}. We fine-tune LXMERT to form visiolinguistic representations based on image and text sequences with adaptive approaches mentioned above. This operation is followed by a classifier that receives the concatenated pooled features of image and text to predict the answer. Fine-tuning is performed on a single P100 GPU with 128 batch size. Optimization is performed with Lookahead~\cite{zhang2019lookahead} with LAMB~\cite{you2019large} as the inner optimizer. Learning rate schedule is regulated by Cyclical LR~\cite{smith2017cyclical}, with base and max learning rates set to $1e-5$ and $1e-4$.

\section{Experimental Findings and Results}
\paragraph{Adaptive span for understanding the complexity of the input sequence}

We demonstrate how learning spans can help in understanding the behavior of individual layers. \autoref{fig:span_results} shows how span varies amongst different attention layers. Studying spans can help us understand which layers are more sensitive to the input sequences encountered during the training process. 

In the case of single modality encoder, spans for self-attention layers for vision and language decrease monotonically, indicating that the learning behavior is somewhat similar, although slopes tell us that the rate of learning is dissimilar. Similar behavior is seen in the cross-modality encoder for language.

 Requiring a larger context size is indicative of the complexity of the sequences. When self-attention attends to both modalities, we observe that the intermediate layers responsible for forming complex representations increase their spans. This observation shows that a more significant span is necessary to attend both modalities jointly. Self-attention also requires a high span when attending to visual features in the cross-modality encoder. This observation shows that visual sequences are perceived as a more complex input to process than a language input in the cross-modality encoder.

\begin{figure*}[t]
%\centering
 % \begin{}{lllll}
% \begin{subfigure}{.5\textwidth}
\minipage{0.3\linewidth}
\centering
    \includegraphics[width=.9\linewidth]{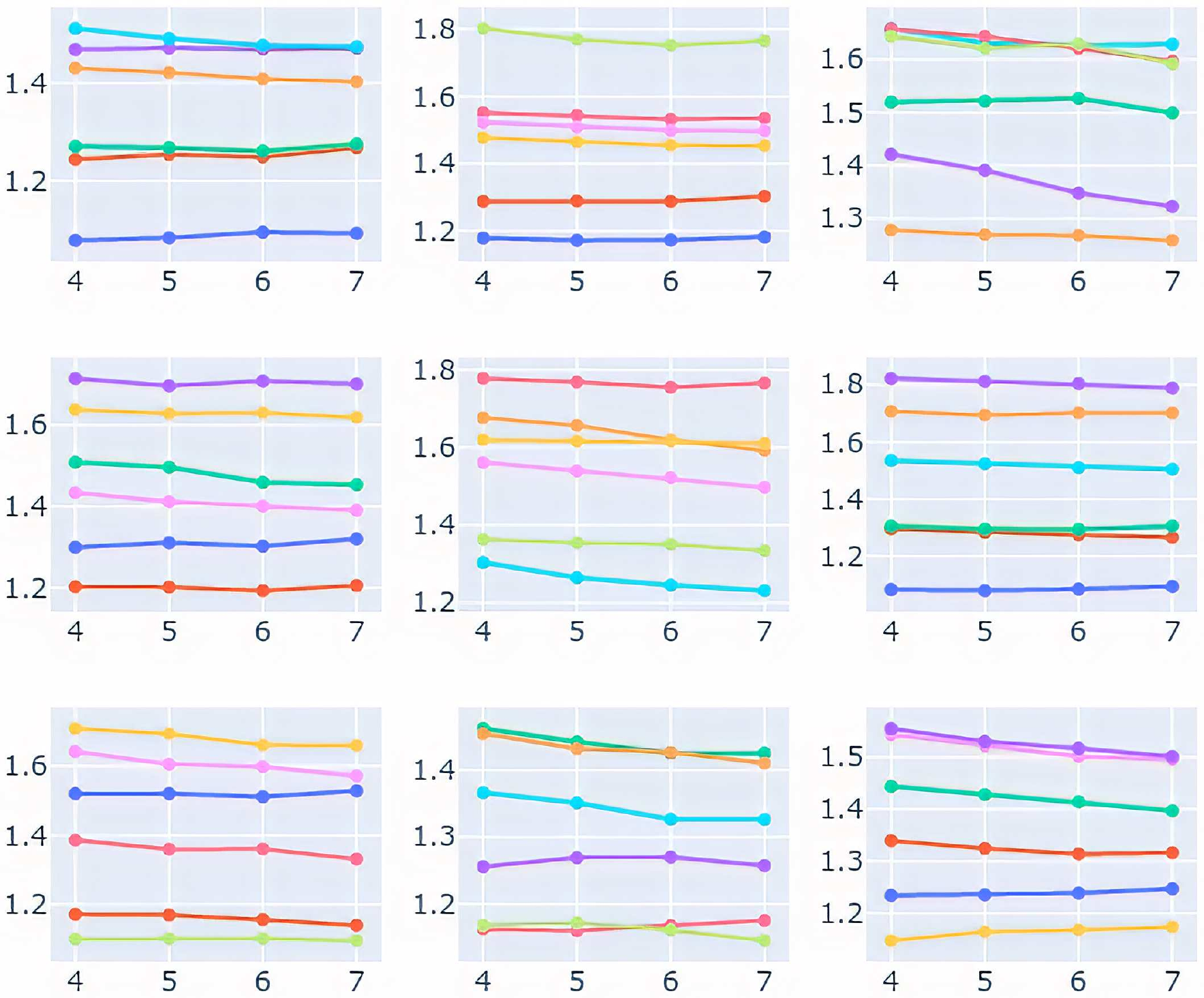}
     \caption*{Language Encoder (9 layers)}
\endminipage\hfill
\minipage{0.3\linewidth}
      \centering
    \includegraphics[width=.9\linewidth]{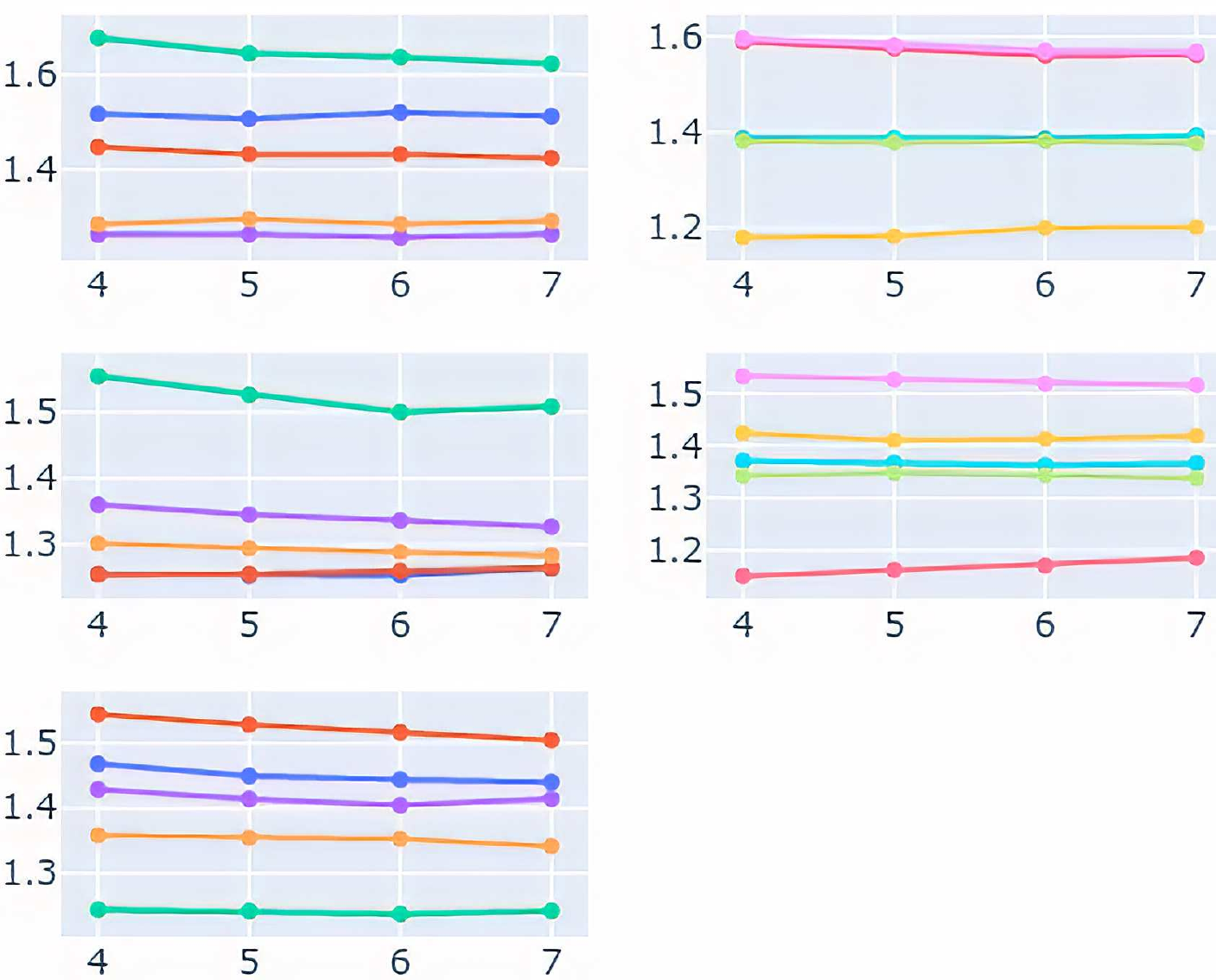}
     \caption*{Cross Modality Encoder (Language) (5 layers)}
\endminipage\hfill
\minipage{0.3\linewidth}
      \centering
    \includegraphics[width=.9\linewidth]{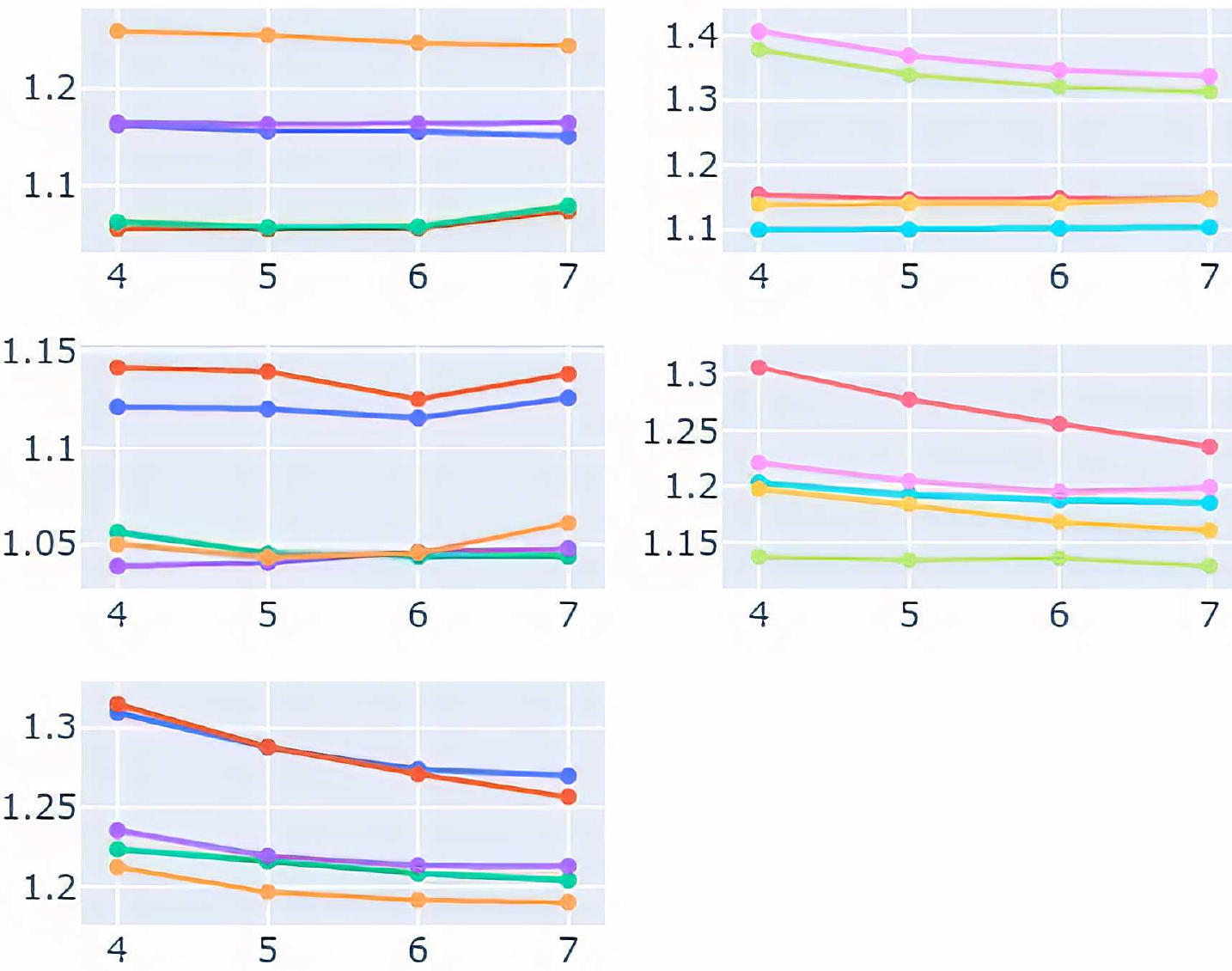}
       \caption*{Cross Modality Encoder for Vision and Language (5 layers)}
\endminipage\hfill
 %   \medskip
\minipage{0.3\linewidth}
      \centering
    \includegraphics[width=.9\linewidth]{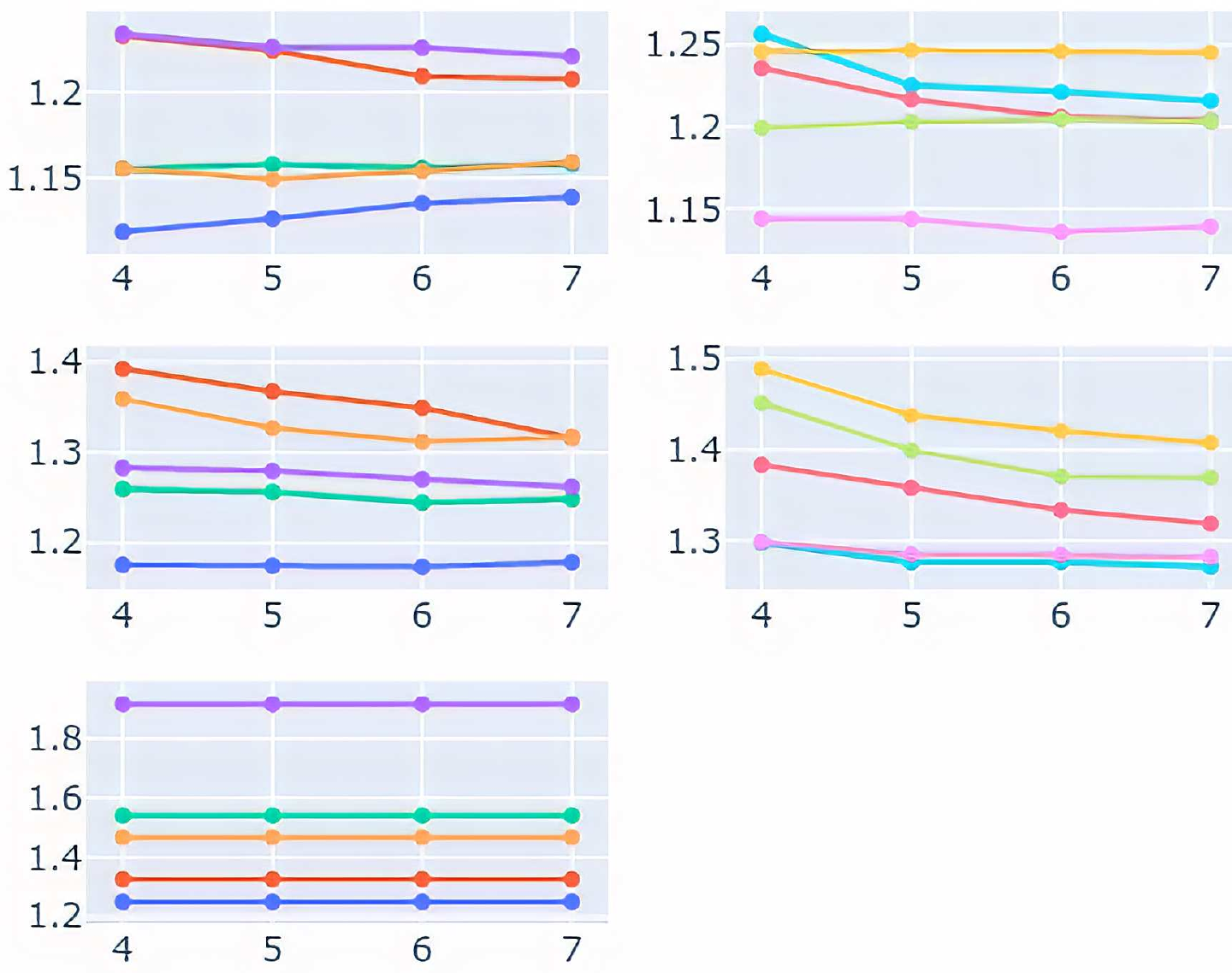}
      \caption*{Cross Modality Encoder for Vision (5 layers)}
\endminipage\hfill
\minipage{0.3\linewidth}
\centering
\includegraphics[width=.9\linewidth]{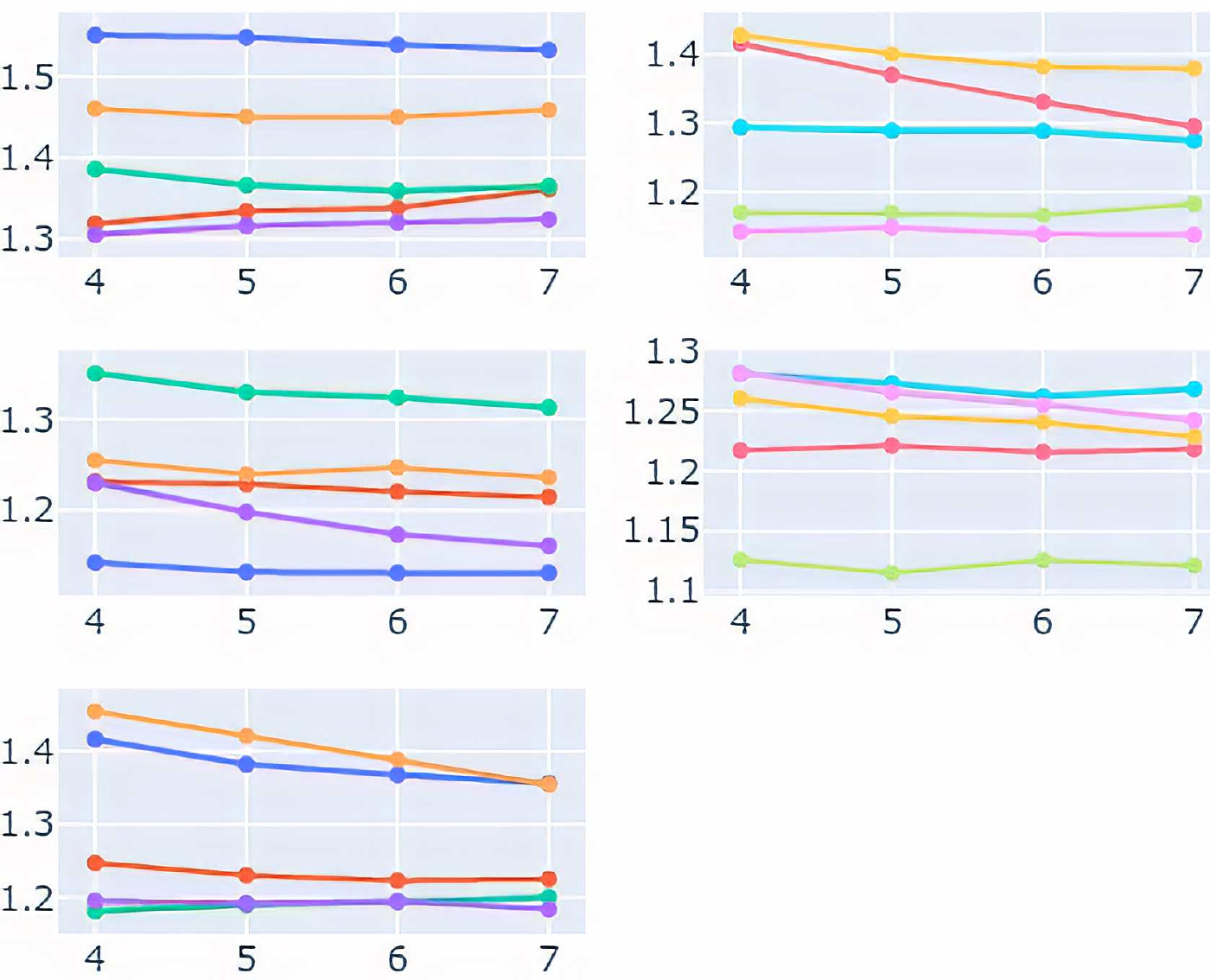}
     \caption*{Vision Encoder (5 layers)}
\endminipage\hfill
 % \end{tabular}
  \vspace{-4mm}
  \caption{Variation of Alpha in Entmax in first six attention heads during an intermediate training stage of 9-5-5 LXMERT model. X and Y axis denote epoch and alpha values, respectively. For simplicity, we only show alpha values for the first six attention heads (12). Color codes denote different attention heads.}
\label{fig: alpha_values}
\end{figure*}

\paragraph{Determining sparsity preferences for vision and language modality with $\alpha$}
The value of $\alpha$ determines if the head is favoring sparse or dense attention weight distribution. For dealing with language modality, self-attention favors mostly sparse mapping of attention weights in intermediate layers. Similar behavior is observed inside cross-modality encoder as well. This observation shows that language modality benefits from sparse weights being assigned as attention distribution. The value of $\alpha$ is restricted below 1.5 for processing visual inputs. When vision modality is involved, heads that preferred sparse mapping initially are converging towards denser mapping, indicating that this representation of attention weights is preferred. We also observe that when both modalities are involved, the network prefers, even more, denser weight distribution. This observation shows that vision modality is given more preference (partly due to perceived complexity) over language inputs to process the sequence.
\autoref{fig: alpha_values} shows variation of $\alpha$ values as training progresses.

\begin{figure}
    \includegraphics[width=0.4\textwidth]{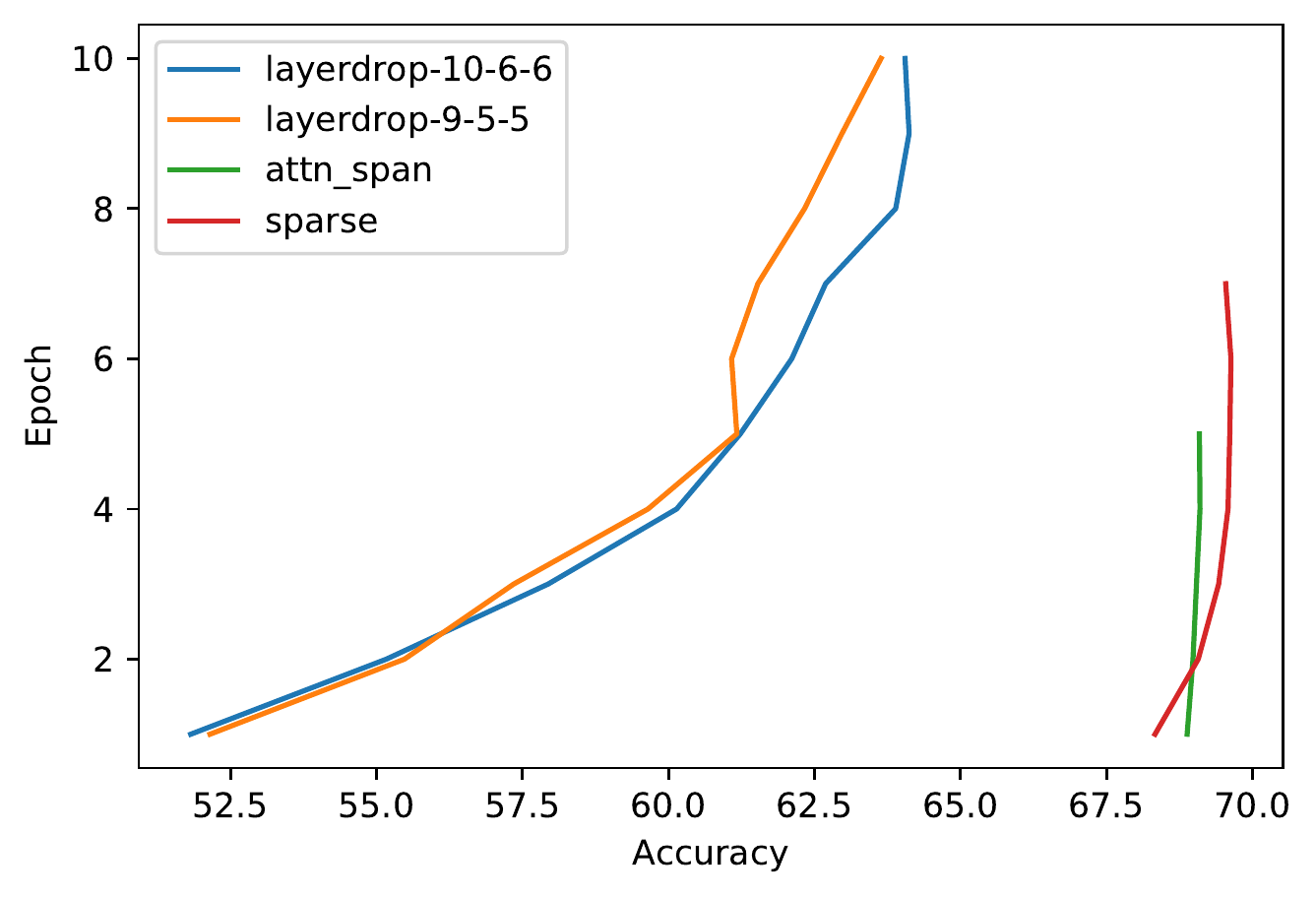}
  \caption{Regularization effect of layerdrop}
  \label{fig:comp_accu}
\end{figure}

\paragraph{Regularization effect of Layerdrop}
We consider two configurations of the model. The first one has 10 language, 6 vision, and 6 cross-modality layers with drop rate ($p$) set to 1 layer. In this case, the number of parameters is more, but the FLOPS is equivalent to the standard 9-5-5 baseline configuration. The later one has the 9-5-5 configuration with $p$ set to 1. This rate causes a FLOP reduction of 17.54\%. It is observed that layerdrop requires $\sim$3.5x more compute runtime for convergence during training. A possible explanation can be that additional training aids in forming a consolidated understanding of multi-modal representations. Even after ensuring the convergence of the model, a strong regularization effect (with a minimum value of p) prevents the network from achieving performance that is close enough with the mentioned adaptive methods with an equivalent number of parameters being used training. \autoref {fig:comp_accu} and \autoref{tab:ablation_results} shows this noted observations. 

\begin{figure*}[t]
\centering
  \begin{tabular}{lccc}
    \includegraphics[width=.2\linewidth]{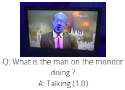}&
    \includegraphics[width=.2\linewidth]{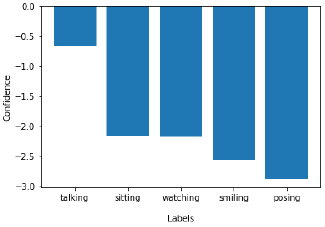}&
    \includegraphics[width=.2\linewidth]{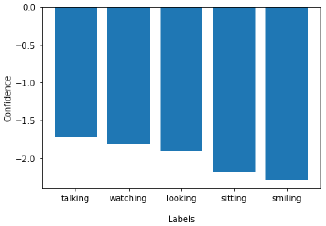}
    \includegraphics[width=.2\linewidth]{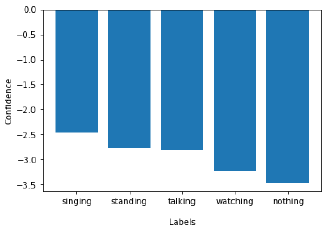}
  \end{tabular}
  \vspace{-4mm}
  \caption{Top 5 confidence scores of an example input sequence \textbf{Left:} Adaptive Entmax \textbf{Center:} Adaptive Attention Span
  \textbf{Right:} 10-6-6 config with Layerdrop (p=1). Zoom in to see scores and labels.}
\label{fig:conf_scores}
\end{figure*}

\paragraph{Quantitative Analysis}
In this section,  \autoref{tab:vqa_results} compares the adaptive approaches with the baseline model and other state-of-the-art models, which rely upon standard softmax attention mechanism. We notice that these approaches achieve near close performance as standard attention mechanisms by being computationally efficient. The results are reported without any hyperparameter tuning. 

\begin{table}[t]
\centering
\setlength{\tabcolsep}{2.5pt}
\begin{tabular}{lcc}
  \toprule
  Model & \ test-dev & test-std \\
  \midrule
  BUTD \cite{anderson2018bottom} & 65.32 & 65.67 \\
  \midrule
  ViLBERT \cite{lu2019vilbert} & 70.55 & 70.92\\
  VLBERT \cite{su2019vl} & 71.16 & - \\
  VisualBERT \cite{li2019visualbert}& 70.80 & 71.00 \\
  UNITER \cite{chen2019uniter} & 72.27 & 72.46 \\
  \midrule 
 \multicolumn{3}{l}{\emph{LXMERT ~\cite{tan2019lxmert}}}\\
  w/ softmax & 72.42 & 72.54 \\
  w/ Adaptive Attention Span & 71.62 & 71.72 \\
  w/ Adaptive Sparse & 71.73 & 71.97 \\
  w/ Layerdrop (10-6-6) (p=1) & 66.4 &  66.72\\
  \bottomrule
\end{tabular}
  \caption{
    Comparison to the state-of-the-art methods with adaptive approaches on the VQA dataset.
  }
\label{tab:vqa_results}
\end{table}

\paragraph{Qualitative Analysis}
In this section, we analyze the confidence scores on complex examples to better understand the network's predictions.
We usually take the class with maximum confidence, but analyzing confidence scores of other classes can help us learn about what the network is learning about the similarity of different tasks in the image. \autoref{fig:conf_scores} shows confidence scores on an example input. We observe that entmax aids in forming a consolidated understanding of contrastive features. In most cases, the top 5 confidence scores include predictions present in the ground truth. Due to sparse mapping, the network makes strong, confident predictions about one label. When trained with an adaptive attention span, the network sometimes seems unsure about the correct label as expected from softmax behavior. It works well when a high probability is assigned to one label in the ground truth. We did not observe comparable performance from Layerdrop. In this example, the right answer is assigned a deficient score. The network does not seem to learn distinguishing features from similar classes properly.

\section{Ablation Analysis}
 We normalize attention scores with entmax instead of softmax before applying the masking function to use both adaptive attention span and sparse attention weights mapping. It is evident from \autoref{tab:ablation_results} that the adaptive span works better with the denser representation of attention weights to perform optimally. The effect of soft masking function is reduced when used with a sparse mapping function. We evaluate the layerdrop method with two configurations of the network 9-5-5 (language, vision, and cross-modality layers) and 10-6-6  with $p=1$. From \autoref{tab:ablation_results}, we see that the shallower network performs better than the deeper-layered model. This observation shows that there is a specific threshold drop rate up until which layerdrop helps. It is plausible that this type of regularization is favorable in deeper networks.

\begin{table}[t]
\centering
\setlength{\tabcolsep}{2.5pt}
\begin{tabular}{lcc}
  \toprule
  Model & \ test-dev & test-std \\
  \midrule
 \multicolumn{3}{l}{\emph{LXMERT ~\cite{tan2019lxmert}}}\\
  w/ Attention Span and Entmax & 63.07 & 63.33 \\
  Default (10-6-6) & 66.35 &  66.57\\
  w/ Layerdrop (9-5-5) (p=1) & 66.51 & 66.81 \\
  \bottomrule
\end{tabular}
\caption{Ablation study for Adaptive approaches}
\label{tab:ablation_results}
\end{table}

\section{Conclusion}
While attention-based approaches are becoming universal, computationally efficient ways must be favored for broader adoption of provided pre-trained models on low resource hardware. Adaptive methods can significantly reduce the cost incurred to train such models and carbon footprints. In this work, we extend adaptive approaches to Visiolinguistic tasks to understand more about attention and adaptive mechanisms. While the empirical results are encouraging, important future work includes explorations of higher efficient adaptive and sparse mechanisms that can significantly cause FLOPS and parameter reduction with minimal loss in performance.

\bibliography{acl2020}
\bibliographystyle{acl_natbib}

\end{document}